**Research Article**

# Humanoid Robot-Application and Influence

**Avishek Choudhury[1,2,3], Huiyang Li[4], Christopher M Greene[4], Sunanda Perumalla[5]**

[1]Applied Data Science, Syracuse University, New York, USA

[2]Process Improvement, UnityPoint Health, Iowa, USA

[3]Systems Engineering, Stevens Institute of Technology, New Jersey, USA

[4]Systems Science and Industrial Engineering, Binghamton University, NY, USA

[5]Computer Science, Texas Tech University, Lubbock, USA

[*]**Corresponding Author:** Avishek Choudhury, Research Assistant, Sociotechnical Engineering, School of Systems and Enterprises, Stevens Institute of Technology, New Jersey, USA; E-mail: achoud02@syr.edu/achoudh7@stevens.edu; Tel: +1 (515) 608-0777

## Abstract

Application of humanoid robots has been common in the field of healthcare and education. It has been recurrently used to improve social behavior and mollify distress level among children with autism, cancer and cerebral palsy. This article discusses the same from a human factors' perspective. It shows how people of different age and gender have a different opinion towards the application and acceptance of humanoid robots. Additionally, this article highlights the influence of cerebral condition and social interaction on a user's behavior and attitude towards humanoid robots. Our study performed a literature review and found that: (a) children and elderly individuals prefer humanoid robots due to inactive social interaction; (b) The deterministic behavior of humanoid robots can be acknowledged to improve social behavior of autistic children; (c) Trust on humanoid robots is highly driven by its application and a user's age, gender, and social life.

## 1. Introduction

Humanoid robots have been assisting humankind in various capacities. They have been broadly used in the field of Healthcare, Education, and Entertainment. A humanoid robot is a robot that not only resembles the human's physical attributes, especially one head, a torso, and two arms, but also can communicate with humans, take orders from its user, and perform limited activities. Most humanoid robots are equipped with sensors, actuators, cameras, and speakers. These robots are typically preprogrammed for specific actions or have the flexibility to be programmed according to the user requirement. Generally, humanoid robots are designed according to their intended application.

Based on applications, humanoid robots can be broadly categorized into Healthcare, Educational and Social humanoid robot.

Healthcare humanoid robots are designed and used by individuals at home or healthcare centers to treat and improve their medical conditions. These robots either require a human controller or are preprogrammed to assist patients.

Educational humanoid robots are primarily designed and equipped for students and are used in education centers or home to improve education quality and increase involvement in studies. These robots are typically but not always manually controlled robots.

Social humanoid robots are used by individuals or organizations to help and assist people in their daily life activities. These robots are commonly preprogrammed to perform mundane tasks and are also known as assistive robots.

### 1.1 Research questions

Manufacturing, healthcare, hospitality, education, and many other fields use robots in some form; Moreover, the application of robots in the military, construction, and research is well-established. A humanoid robot is a specific robot that is in its developing phase. Humanoid robots can perform several human-like physical activities; however, its effectiveness, especially in the field of healthcare, education and social, is a concern. In this study we are focusing on the following questions:

- How people of different age and gender has a different opinion towards the application and acceptance of humanoid robots?
- Can cerebral condition and social interaction of individual influences her or his behavior and attitude towards humanoid robots?

This paper conducts a systematic literature review to answer the above concerns.

### 1.2 Literature review

Our study conducts a systemic literature review to analyze the impact and application of humanoid robots undertaken in the last ten years. The search for this review was limited to ACM Digital Library, ASME Digital Collection, BIOSIS Citation Index, CINDAS Microelectronics Packaging Material Database, Cite Seer, Computer Database, Emerald Library, Energy & Power Source, Engineering Village, IEEE Xplore, MEDLINE, OSA Publishing, PubMed, Safari Books Online, Science Direct, Sci Finder, SPIE Digital Library and Springer Database. The initial search result using the keyword "humanoid robot" gave 12,261 results that included books, articles, conference proceedings, newspaper articles, dissertations, retracted papers, technical reports, audiovisuals, government documents, statistical datasets, and images.

### 1.3 Selection/exclusion criteria

Our focus was restricted to papers published between 2008 and 2018. Further constraints such as material type and language were set. Papers published in the English language were listed, this left us with 3,392 papers to investigate.

**1.3.1 Title based selection:** The first scan of these papers was done by reading the title only. Papers focusing on human-robot interaction and application of humanoid robots were considered which left us with 368 peer-reviewed articles.

**1.3.2 Abstract and finding final based selection:** The selected papers were again scanned to make a definitive list of documents. This was done by reading the abstract of each article. We preferred to any paper that has any connection with education, healthcare, and social. Fifty-eight papers were shortlisted for this study.

**1.3.3 Findings:** Humanoid robots have been used in the field of healthcare and education. The majority of the study involved minors and senior citizens; However, economic feasibility was not tested in any study included in this paper.

The following section briefly discusses the influence of age, gender, and other experimental setups on human behavior toward humanoid robots and their application specifically in the field of healthcare and education.

**1.3.4 Healthcare Humanoid Robot:** Healthcare practitioner and benefactors have appreciated the advantage of advanced surgical robots. However, our study highlights the application of humanoid robots and their roles in healthcare. In addition to surgical robots, healthcare humanoid robots have been successfully helping people in disease management, pain relief, pediatric healthcare assistant, and physical therapy. The role of healthcare robots can be broadly classified into the clinical and non-clinical application.

**1.3.5 Clinical application:** In Clinical setting, humanoid robots have been used to assist patients with cerebral palsy [1], and pediatric cancer [2]. To study the influence of human-robot interaction two children of age 9 and 13 with cerebral palsy were exposed to NAO robot under four different interactive situations. The experiment aimed at improving patient coordination, truncal balance and motor function [1]. The first interaction was a general introduction round where children and robot verbally communicated with each other. In this situation, the subject had a tough time understanding the robot and required the help of a therapist; thus, it increased positive interaction between the subjects, humanoid-robot, and the therapist. This was aimed to enhance a child's social adaptability [1]. The second, third, and fourth interaction session was an imitation round which aimed at improving the lower leg balance and function. In this setup, the children had to imitate the movements of the humanoid robot by lifting one leg and kicking a ball [1]. No improvement was observed during this setup; however, the children developed a positive interaction with the humanoid robot [1].

The essential responsibilities of the robot in the clinical healthcare domain are mollification of distress [3], remote monitoring [4], and interacting with the patient [1]–[5]. To measure the impact of a humanoid robot in pain and distress mollification, "Face Pain Scale-Revised" [3] approach was taken during the vaccination of children in a clinical setup. The pain experienced by the children during an injection shot was measured through their facial expression and behavior such as crying and muscle tension. Children felt more pain during vaccination in the absence of a robot in the clinic [3] than in their presence. Moreover, studies on the effect of a humanoid robot on anger, anxiety, and depression level have also been significant. To study the effect of humanoid robots on anger, and anxiety, "social robot-assisted therapy" [2] and psychotherapy were compared by giving individual psychotherapy and psychotherapy using a humanoid robot to a different group of children with cancer. The group assisted by the humanoid robot had eight sessions in which the robot played different roles such as a doctor, chemotherapist, nurse, cook, ill kid and other. In these sessions, the humanoid robot interacted with the children and explained to them the role of each character in a story form to reduce their anger, depression, and anxiety. Post the experiment a questionnaire was used to assess anger, fear, and depression level of the children. It was observed that the robot-assisted group had lower anxiety, depression, and violence than that of the controlled group. Thus, humanoid robots were successful in minimizing anger, anxiety, and depression [2] among cancer patients. Humanoid robots also enhanced joint attention between the patient and the therapist [6].

The patients treated using humanoid robots were mostly elderly individuals and adolescents around the age of 85 years [4]; 5 to 6 years [3]; 7 to 12 years [2]; 5 to 14 years [6]; 9 & 13 years [1]; Moreover, most of the study required a supervisor or controller to monitor and control the humanoid robot.

**1.3.6 Non-clinical application:** Non-Clinical healthcare have significantly contributed to autism management followed by diabetes management [7] by performing activities such as playing games [8]–[10], greeting, singing, dancing, hand movement, blinking, interacting with the patients [2], [3], [11]–[13]. Robots also measured blood pressure [14]–[16] and asked questions, played a quiz with the patients [7][17], monitored and helped patients with medical assistance [4].

The effect of using robots in autism management has been highly effective and appreciated. Humanoid robots can be used to foster social and behavioral skills within autistic children [18], thus, can improve patient's autistic behavior [19]–[21].

Gaze is a crucial medium that enables social communication. It also affects acceptance, preference, and obedience among human beings [19]; However, excessive gaze might impose a threat, superiority, and anger [19]. Fifty-two University students participated in an experiment in which they had to engage in the "shell game" [19] with a humanoid robot. The game consisted of three different levels of difficulty. A mixed 3 X 3 design was employed to study the behavior of the subjects at Averted gaze, constant gaze and situational gaze [19] for the easy, medium and hard difficulty level of the game. Here, the independent variable was the three levels of gaze and game difficulty level — averted gaze in which the robot never looked at the participant, constant gaze during which the robot continuously seemed at the participant and situational gaze when the robot looked at the participant only when he or she gave a wrong answer. It was observed that with an increase in difficulty participant's trust towards the humanoid robot increased.

Use of humanoid robot had a significant influence on communication, social behavior and joint attention of autistic patients [22], [23] but did not influence any collaborative behavior among patients [10]; However, playing with human adult enhanced collaboration among patients [24].

A therapy session in a school playroom was set up to study the effect on joint attention. Students in this experiment used a humanoid robot, and the interaction was recorded. In this study, the robot directed students to perform physical activities such as touching head or look towards the window. The study measured the number of times the participants responded correctly to the robots. The experiment was run under two conditions, with prompting and without prompting. Improvement in joint attention initiation and response were noticed; however, the relevance and contribution of prompting were unclear.

Humanoid robots helped patients to learn more about autism management [8], [9]. The ability of the humanoid robot to evoke human-human interaction along with its impact on learning was analyzed in a study conducted by Costa in 2015. The experiment was classifieds into four sections: "familiarization, pre-test, practice, and post-test phase" [8]. In this experiment, the participants had to identify their different body parts as per asked and directed by the robot. In this study, it was observed that the participant's response increased in the "post-test phase" [8], [9] and; thus, the learning ability of the participant were increased after therapy using a robot. The study, thus, showed the effectiveness of a humanoid robot in the domain of child education and healthcare,

Humanoid robots have also enhanced diabetes and diabetes management [7]. The benefactors of the humanoid robots in autism and diabetes management were mostly children [24], [25] of age 6-7 years [23], 6-9 years [8], [9], 8-12 years [7], 5-10 years [26], [27], 6-8 years [24], 7-12 years [7], 5-13 years [10] and 7-13 years [13].

**1.3.7 Education Humanoid Robot:** Use of computer and e-learning in the field of education have been performing well and have successfully increased the accessibility to education worldwide. However, the recent trend in education domain is towards the application of humanoid robots. Humanoid robots are now on the verge of becoming an essential component in the field of education as these robots can reason and analyze situations logically to support human learning and are also better than computer agent [28] and more engaging than the virtual agent [29]. Comparison between a projected robot, a collocated robot, and an on-screen agent has been a relevant concern in the domain of education and e-learning. To compare the impact of a computer agent, on-screen projection, onscreen projection of a robot and a physical robot on the social behavior such as engagement, disclosure, influence, memory, attitude, and others were measured to find that collocated or physical presence of robot-enhanced participant's involvement [28] with the subject. However, it did not affect social behavior [28]; Moreover, there was no significant difference found between onscreen robot, and a collocated robot [28]. Unlike other studies, this showed that learning ability was minimum using an arranged robot [28]. Humanoid robots have been known for teaching language[30] [17], hands-on engineering [31], nutrition [32], mathematics [33], general science [34] as well as helps students in learning spellings, storytelling [35] and participate in memory games. Robots have been performing the role of a teaching assistant[36], [37][38] and games partner of children [39].

In most of the studies, humanoid robots were used along with a human teacher or a controller. The educational humanoid robots have been used for various sections of education and have addressed wide range of students such as preschool kids [33] [35] [39], primary school kids [17], [33], [35], [38], [40], [41] junior high school students [36] and undergraduate engineering students [42]. Students responded positively to the robots. Positive effect on learning [33] was observed along with higher participation [34]. Increase in a student's creativity [43], curiosity, knowledge, and recall rate [29], [44] were observed.

**1.3.8 Socially assistive robot:** Social robots or socially assistive robots (SAR) are known as assistive robots, and their application is burgeoning especially among elderly people and hospitality industries. The social robot assists human beings in their daily life and replaces human activities at hospitality industries. At the domestic level, social robots have been doing well. Typically, older adults and autistic patients prefer and are benefiting from humanoid robots. These assistive robots act as a companion [45], [46] for both children and elderly individuals. Among children, social robots have been known for their ability to entertain [24] and play games [46], [47]. Importantly, autistic children prefer to spend more time with a robot [13] since robots are more predictive and less intimidating than human beings [48].

In this domain, most of the study emphasized the effect of age, gender and appearance of the robot on acceptance, social behavior of the user, and trust towards the humanoid robot. Children and elderly users have a diverse opinion about the appearance of a robot. Some preferred humanoid whereas, some liked machine-like appearance [11]. Since humanoid robots look like a human, they are more relatable and a better fit for companionship [11][49]. According to a literature review, people relate male humanoid robots as more intelligent in decision making, whereas; female robots were perceived to be good at nursing and caring activities [11].

Humanoid robots can also evoke the feeling of care and enhance awareness of social behavior [47]. Consecutively, according to a survey and interview conducted by Broadbent in 2009 older people opted for robot staff to help them in daily life activities such as making a phone call, control appliances, remind medications and appointments [49]. In the study by Broadbent in 2009, several pictures of different types of robots were shown to the workers and residents of a nursing home. Most individuals preferred a humanoid robot with all human-like physical features. Even the size of the robot impacted their perception. Elderly users opted for a medium sized robot with light, bright color; However, there was no influence of the robot's gender of the users [49]. The ability of a social robot to detect falls and serious medical condition of the user was highly appreciated; Moreover, assistive robots also play an important role in elderly care, they remotely observe the user and communicate with their care providers, thus, further reducing nurse workload [50]. Table 1 below summarizes the main findings from the literature review.

| Author | Year | Finding | Robot | Method | Participants | Domain |
|---|---|---|---|---|---|---|
| Taheri, Meghdari | 2018 | Decrease in Autism severity; Improvement in social behavior & participation. | NAO | Games with the robot; questionnaire; interviewing parents | n=6; age=6 to 7 yrs. | Healthcare |
| Charron, Lewis | 2017 | Improved communication skills | NAO | Speech therapy session with a robot | n=1; age= 8yrs. | Healthcare |
| Miyachi, Iga | 2017 | The paper certified that humanoid robots could substitute caregivers | PALRO | Experiment (recreations and health gymnastic activities with robots), feedback from participants | n=13/38; age: 70 to 87 yrs./ 65+ yrs. | Healthcare |
| Bakster, Ashurst | 2017 | Personalized robots are accepted more than a regular robot; Personalized robots generate higher learning | NAO | Experiment and Questionnaire | n=59; age: 3 yrs. | Education |
| **Erich, Hirokawa** | 2017 | Most used Robot is NAO in the field of Teaching, Assisting, Playing, and Instructing | KASPAR; NAO; BANDIT; ROBOT; ROBOVIE R3 | Literature review | (----) | Healthcare, Education |
| **Stanton, Stevens** | 2017 | Females did not trust robots when they steadily gazed at them, but they trusted the robot for situational gaze, participants trusted the robot as game difficulty level increased, participants with low confidence about their answer, believed robots more. | NAO | Experiment, game at three levels of difficulty | n=52; age: 22.5 yrs. (Mean) | Social |
| **Thelma, Silvervarg** | 2017 | People find robots to take undesirable actions more than human; user associate negative behavior to robots, users believe robots do everything intentionally. | Ellis | Questionnaire, | n=90; age:24 yrs. (mean) | Social |
| **Vandemeulebroucke** | 2017 | Participants gave more attention to SAR's ease of use, user like interacting with the robot, Human looking | | Literature review, wizard of Oz, Thematic synthesis by Thomas and Harden in 2008 | n= 23 studies; age: 65 yrs. (Mean) | Social |

| | | | | | |
|---|---|---|---|---|---|
| | | SAR were preferred | | | |
| **Baxter** | 2017 | Children showed significantly increased learning in personalized condition | NAO, Sandtray touch screen | classroom experiment, robot taught three subjects, picture recall, math, and spelling | n=59; age:7 to 8 yrs. | Education |
| **Henkemans, Bierman** | 2017 | Improved health literacy in children. The learning activity was entertaining, engaging and motivating. | NAO (personal and neutral) | Experiment: introduction, quiz | n=27; age: 7 to 12 yrs. | Healthcare, Education |
| **Llamas, Conde** | 2017 | Students liked the robot more than human teachers. | Baxter | Experiment; observing the reaction and behavior of the students while interacting with the robot. | n=210; age: 6 to 16 yrs. | Education |
| **Alemi, Ghanbarzadeh** | 2016 | A robot can be used as an assistant in cancer treatment. Humanoid robot was found to be useful in teaching children about their afflictions. Robots instructed the children about methods to confront their distress level. | NAO | Children interacted with robot, Children responses such as anger, and depression was measured using questionnaire | n=11; age: 7 to 12 yrs. | Healthcare |

| Yun, Kim | 2016 | Robots assisted behavior intervention system was developed that is capable of facilitating social skills for children with Autism | iRobiQ | Experiment; | (---) | Healthcare |
|---|---|---|---|---|---|---|
| Gaudiello, Zibetti | 2016 | Users trust robot's functional knowledge (weight, color, height, etc.); Users did not trust robots, social knowledge (most important subject, an important object, etc.); Collaborative interaction with robot did not affect the users' trust | iCub | Experiment, Questionnaire | n=56; age: 19 to 65 yrs. | Social |
| Pennisi | 2016 | NAO is the most widely used the robot in Autism therapy, participants with Autism had better performance in robotic condition than in human condition | NAO, KASPAR, Pleo, Tito, and others | A systematic review, (electronic database search) | n= 29 studies | Healthcare |
| Yi, Knabe | 2016 | Using humanoid robot motivates college students to be interested in engineering, hands-on experience with robot helps in better learning | Darwin-HP | Hands-on experience with the robot. modified design of the robot with the help of a supervisor | n=65; age: undergraduates | Education |
| Hashim | 2016 | Teachers demanded to see other school using humanoids in the classroom before they accept one, Robots must be personalized especially for autism patients. | NAO | Literature review, interviews with teachers & parents | age: Parents and teachers | Healthcare |
| Huijnen | 2016 | NAO is used in the highest number of articles, and Robots focus on very set of objectives. However, ASD treatment requires to cover a wide range of domains | NAO, and other robots | the focus group, and systematic literature study | n =53; age: adults | Healthcare |
| Rosi, Dall'Asta | 2016 | The robot's presence along with the teacher did not affect the learning improvement. | NAO | Experiment: robot intervention with the teacher; Questionnaire | n=112; age:8 to 19 yrs. | Education |
| Aziz, Moganan | 2015 | Children got more involved when robot talks and makes hand gestures. | NAO | Experiment: interaction with the robot. (greeting, singing, etc.); Observation using Kansei checklist | n=3; age: children | Healthcare |
| Rahman, Hanapiah | 2015 | Robot's movement encourage imitation learning, socializing and motivate the children | NAO | Experiment: Interactive session with robot | n=2; age: 9 & 13 yrs. | Healthcare |
| Malik, Yussof | 2015 | The robot improved treatment efficiency by initiating joint attention between child and therapist. | NAO | Experiment: Interaction with the robot. | Age: 5 to 14 yrs. | Healthcare |
| **Mann, MacDonald** | 2015 | Robots are more trusted than tablets; Appearance of robots determines the user's responses to healthcare interactions; Increased speech with robots; More emotion and relaxation with robots; Participants thought that robot was accurate in measuring blood pressure than a tablet | Yujin Robot's iRobiQ, ASUS Google Nexus 7 Tablet | Interaction with robot and questionnaire | n=65; age: 19 to 62 yrs. | Healthcare |
| **Alemi** | 2015 | The robot can be used as assistant in cancer treatment, and the Humanoid robot was found to be useful in teaching children about their afflictions, Robots instructed the children about methods to confront their distress level. | NAO | Children interacted with robot, Children responses such as anger, and depression was measured using questionnaire | 7-12 yrs.; 11 children | Healthcare |
| **Barakova** | 2015 | Interaction with robots induced more creativity among children. | NAO | 12 sessions with the robot and student LEGO designing experiment | n=6; age:8 to 12 yrs. | Education |

| Shiomi, Kanda | 2015 | It encouraged the children to ask about science by initiating conversations about class topics. However, the robot did not increase any curiosity in the subject. Some students asked more questions to the robot. | Robovie | Experiment | n=144; age: 4th to 6th-grade student | Education |
| --- | --- | --- | --- | --- | --- | --- |
| Li, Lizilcec | 2015 | Participants who saw the inset video of the actual lecture replaced by an animated human lecturer recalled less information than those who saw the recording of the human lecturer. However, when the actual lecturer was replaced with a social robot, knowledge recall was higher with an animated robot than a recording of a real robot. | NAO | Experiment: online video course | n=40; age: 18 to 25 yrs. | Education |
| Ioannou, Andreou | 2015 | Kids can easily interact and play with the robot. Children also took care of the robot when he falls. | NAO | Experiment: Playing with NAO | n=4; age:3 to 5 yrs. | Social |

| Huskens, Palmen | 2015 | The robot intervention had no significant effect on the children. It did not improve the collaborative behaviors of children | NAO | Experiment: play Lego with robot | n= 6; age: 5 to 13 yrs. | Healthcare |
|---|---|---|---|---|---|---|
| Costa, Lehman | 2014 | Children did not lose interest throughout the session. Games with robots increased children's learning. | KASPAR | Experiment: Familiarization with robot-> practice task -> perform | n=8; age: 6 to 9 yrs. | Healthcare |
| Kachouie | 2014 | SAR enhances the well-being of seniors and minimizes nurse's workload. | AIBO, Bandit, Healthbot, iCat, Ifbot, and other | Systemic review (Cochrane Handbook for Systematic Reviews of Interventions) | n= 86 studies | Social |
| Boboc | 2014 | Robots can be used to attract students to educational institutes | NAO | The robot as a tour guide | (---) | Education |
| Freidin, Belokopytov | 2014 | Teachers are ready to accept humanoid robots to serve as an interactive tool in education, and Teachers believed robots do not evoke negative vibes | NAO | UTAUT and TAM model to check acceptance of robot. The questionnaire, followed by an experiment where participants interacted with the robot | age: preschool & elementary school teachers | Education |
| Broadbent, Kumar | 2013 | A robot with a human-like face is preferred, trusted and considered to be social | People-bot Healthcare robot (face | Experiment: Robot measured user's blood pressure then user rated their experience with the robot | n=30; age: 22 yrs. (mean) | Healthcare |
| | | | displayed on a screen) | | | |
| **Henkemans, Bierman** | 2013 | Children loved robot. They talked more to the robot. They learned more about diabetes. Children mimic the robot. | NAO | Experiment: Robots asked a question to the children | n=5; age: 8 to 12 yrs. | Healthcare |
| **Broadbent, Kumar** | 2013 | A robot with a human-like face is preferred, trusted and considered to be social | People-bot Healthcare robot. | Robot measured user's blood pressure then user rated their experience with the robot | n=30; age: 22 yrs. (Mean) | Healthcare |
| **Henkemans, Bierman** | 2013 | Children loved robot; They talked more to the robot; They learned more about diabetes; Children mimic the robot | NAO | Robots asked a question to the children | n=5; age: 8 to 12 yrs. | Healthcare/ Education |
| **Kamide, Kawabe** | 2013 | Females > Male (robot humanness and familiarity); Middle age females ranked utility to be highest, and young age ranked it least; Old males and females think robots are useful; Adolescent males like robots' utility more than adolescent women. | Robovie, wakamaru, enon, ASIMO, HRP2, HRP4C | Experiment, introduction to robots | n=900; age: 10 to 70 yrs. | Social |
| **Beran** | 2013 | Interaction with robots during flu vaccination resulted in less pain and distress | NAO | robot interacted with the child while the nurse gave a vaccination, robot gave a high-five, introduction and talked about the child's interest | n=57; age:5 to 6 yrs. | Healthcare |
| **Freidin** | 2013 | Children reacted positively with the robot, paid high attention, showed a high degree of enjoyment. | Kindergarten Social assistive robot | robot playing with children, | n=11; age: kindergarten students | Education |
| **Kim, Suzuki** | 2013 | No difference in engagement; While playing with the robot, people gazed at the robot more than on the table | Genie | Experiment: playing the game first with human, then with the robot.; Questionnaire | n=10; age: 22 to 29 yrs. | Social |
| **Wainer, Dautenhan** | 2013 | Children with Autism were more interested and entertained by a robot partner, but the show more collaborative action while playing with a human adult. | KASPAR | Experiment | n=6; age:6 to 8 yrs. | Healthcare/ Social |
| **Cabibihan, Javed** | 2013 | The paper categorizes the robots as a diagnostic agent, playmate, eliciting behavior agent, social mediator, social actor, personal therapist. | | Literature Survey | (----) | Healthcare |
| **Beran, Serrano** | 2013 | Parents wanted robots beside their children while flu vaccination was given. The child smiled more when the robot was present. The child more memory of the robot than of the needle | NAO | Experiment; Interviewing parents | n=57; age: 4 to 9 yrs. | Healthcare |
| **Freidin** | 2013 | Children enjoyed interacting with the robot. | Kind SAR | Experiment: Robot did physical action and pre-recorded storytelling | n=10; age: 3 to 3.6 yrs. | Education |
| **Wood, Dautenhan** | 2013 | Interview using robot lasted longer; Children looked at the robot more than the human; Children were willing to interact with a robot for an interview in the same way they wanted to interact with a human, the information exchange was also similar | KASPAR | Experiment: Interviewing the children (both human and robot interviewed); Questionnaire | n=21; age: 7 to 9 yrs. | Social |
| **Ismail** | 2012 | Less stereotype behavior was observed with robot interaction than in a regular classroom | NAO | Experiment: Interaction between the robot and the children were observed, NAO introduced itself, hand movement, play song, and eye blink; song and hand movement | n=12; age: children | Healthcare |
| **Bacivarov** | 2012 | Humanoid robot resulted in 61%Higher participation, 80%Higher performance, and creative thinking, 0.81 Higher social impact, 100% Course attendance, and zero drops out rate | Robonova | Teaching using actual robot and simulator | n= 15; age: primary school students | Education |
| **Ismail** | 2012 | Less stereotype behavior was observed with robot interaction than in a regular classroom, and Stereotype behavior can be reduced and improved further by the better modules in a robot-based intervention program | NAO | The interaction between the robot and the children were observed, NAO introduced itself, hand movement, play a song and eye blink; song and hand movement | n= 12 | Healthcare |
| **Kahn, Kanda** | 2012 | Children developed a relationship with the robot. The children would like to play with the robot at a free time or when they feel lonely | Robovie | Robot-human interaction, general verbal interaction, played the game, followed by an interview | Healthcare/Social+J90: K103 | Social |
| **Shamsuddin, Yussof** | 2012 | Children showed a decrease in autistic behavior; the robot was able to engage the children. Lower autistic traits were | NAO | Experiment: Interaction with the robot | n=5; age: 7 to 13 yrs. | Healthcare |
| | | observed during the HRI session than compared to the in-class setting. | | | | |
| **Back, Kallio** | 2012 | Humanoid robots can be used as remote monitoring of nursing home residents, and the robot can autonomously perform the checking of the resident's room and provide a caregiver with real-time images and a voice connection to the place. | NAO | Experiment: tested a prototype in three nursing homes | n= 10 to 22; age: 85 yrs. | Healthcare |
| **Chin, Wu** | 2011 | Physical robot helped students understand more; students feel relaxed with the robot than a teacher; Robot and teacher sync was good; All students want robots in their class | IDML TOOL (with the humanoid robot and computer screen) | In-class use of robot and Questionnaire | Primary school students | Education |
| **Chang, Lee** | 2010 | Students enjoyed learning and responded positively with the robot as teaching assistant; the Preferred robot in storytelling mode; the Preferred robot in oral reading mode; the Preferred robot in cheerleader mode Preferred robot in action command mode; Like robot in Q&A mode | Sapien | Experiment | n= 100; primary school students | Education |
| **Han** | 2010 | Robots are more responsive in teaching than e-learning; Robots can initiate learning and active than e-learning; Unlike a computer, robots can have physical interaction with students; Robots can build a relationship with the user; Robots enhances communication between children and parents.; Teacher and student both prefer robots in education | iRobiQ, Tiro, | Literature review | age: Primary school students | Education |
| **Bainbridge, Hart** | 2010 | Participants were more likely to fulfill requests from physically present robot; They allowed more personal space | Nico | Experiment: Interaction with the robot, physically and over video | n=65; age: 24 yrs. (Mean) | Education |
| **Broadbent, Tamagawa** | 2009 | Older people demand robot care provider; People prefer a robot to staff; People prefer robot with a display screen; Robots that can detect fallings, control appliances, remind medication, making phone calls are accepted easily; Users prefer human-like structure in their robot | Hopis and In-Touch Telemedicine Robot | Questionnaire and face to face interview, | age: 60 yrs. | Healthcare and social care |
| **Kuo, Rabindran** | 2009 | Males have a more positive attitude towards robots than females, can be a potential customer; Users demands, more interactive and better voice from robot; Middle-aged and older people responded in the same way | People bot with display monitor (Charles) | Experiment (blood pressure measurement) | n=57; age: 40 to 65yrs. and >65yrs. | Healthcare |
| **Powers, Kiesler** | 2007 | Participants spent the most time with collocated remote and remote robot than computer agent, and women | computer monitor; Agent projected on a | interview with robot and agent, questionnaire | n=113; age: 26 yrs. (mean) | Healthcare |
| | | confessed more than men did, users remembered the most information from the agent, users found robots more useful. | large screen; remote robot projected on a large screen; the collocated robot in the room | | | |
| **Robins, Dautenhahn** | 2004 | The child mimics the robot. The child corrected his/her mistake while mimicking the robot's movement. Eyegaze, closeness to robot, touch, and imitation increased with time | Robota | Longitudinal research, | n=4; age: 5 to 10 yrs. | Healthcare |
| **Robins, Dautenhahn** | 2004 | Children corrected his/her mistake while mimicking the robot's movement, Eye gaze, closeness to robot, touch, and imitation increased with time. | Robots | Longitudinal research | n=4; age: 5 to 10 yrs. | Healthcare |
| **Bruce, Nourbakhsh** | 2002 | People paid more attention when the face was present and tracked them | RWI B21; Human face on a robot display screen | Experiment, (robot asking a question) | (-----) | Social |

**Table 1:** Main Findings.

The table following table 2, lists the typical application and features of commonly used humanoid robots.

| Robots | Application | Control | Robot Action | Actuators | Sensors |
|---|---|---|---|---|---|
| **KASPAR** | Teach children with Autism to identify their body parts | Mixed (manual and autonomous) | Reacting to touch | Motors and speakers | Tactile |
| **KASPAR** | Teach children with Autism | Autonomous | Responding to touch | Motors | Tactile |
| **NAO** | Assist school staff | Manual | Singing, dancing, playing, explaining | Motors and speakers | Microphone |
| **NAO** | Assist older people | Autonomous | Healthcare assisting activities | Motor, Speaker, and Projector | Receiver, Camera, external sensor network |
| **NAO** | Entertainment | Manual | Playing, talking, gesturing | Motors and Speakers | None |
| **NAO** | Education | Mixed | People detection, talking | Motors and Speakers | Camera |
| **NAO** | Deliver letter | Manual | Walking, bowing, waving | Motors | None |
| **NAO** | Train Autism victim's attention skills | Mixed | Asking a question and moving naturally | Motors and Speakers | Camera network |
| **NAO** | Interacting with Autism child | Manual | Sitting, walking, dancing, speaking | Motors, Speakers | None |
| **NAO** | Hospitality (hotel reception) | Autonomous | Looking at guests, reading | Motor, Speaker (Text to Speech) | Kinect |
| **Bandit** | Assist post stroke (healthcare) | Autonomous | Giving instructions, feedback, motivating | Motors, Speakers | Wire Puzzle |
| **Robota** | Interact with Autism child | Manual | Move as per instructed | Motors | None |

| Robovie R3 & NAO | Teach sign language to the child | Mixed | Indicating sign | Motors, LED, Speakers | Kinect, camera, microphone |
| --- | --- | --- | --- | --- | --- |
| NAO | Teach physical exercise to reduce back pain | Manual | Demonstrate activity | Motors | None |

Table 2: Common applications of humanoid robots.

## 2. Discussion

The application of humanoid robots specially NAO [51] has been significant in several domains [52]. It has been successfully implemented in the field of healthcare, education, and social. Figure 1 below shows the application of humanoid robots in three broad domains.

The majority of the research is focused on healthcare. Figure 2 below shows that humanoid robots were used mostly to treat autism (65%) followed by diabetes (15%), cancer (10%) and cerebral palsy (10%). Humanoid robots improved autism severity [12] and enhanced social behavior [22], communication skills among children [22][6]; Moreover, use to humanoid robots improved collaborative behavior [24], learning capacity [8] among autistic children and interaction with robots made them feel entertained [24] and comfortable [27]. Apart from successfully treating autism, humanoid robots yield positive results in educating patients with diabetes management skills [7], minimize stress in pediatric cancer patients [2]. Patients who have cerebral palsy got encouraged [1] by interacting with humanoid robots. Usage of the robot also enhanced treatment efficiency by initiating joint attention between the patient and the therapist [6].

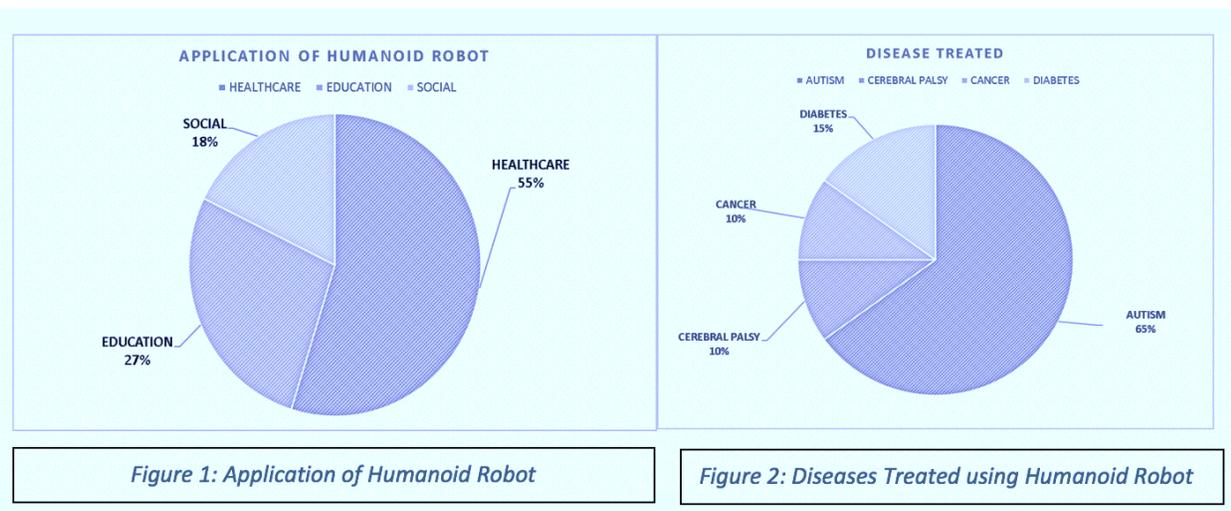

Figure 1: Application of Humanoid Robot     Figure 2: Diseases Treated using Humanoid Robot

Figure 1 shows the proportion of work dedicated to healthcare, education, and social assistance using humanoid robots and highlights [53] the influence of humanoid robots [54] in the field of healthcare, moreover, the potential of a humanoid robot to become a personal healthcare assistant [54] has been projected in a recent study [55]; however, some limitations exist especially to treat autistic children. To better manage autism, humanoid robots must have a diverse set of objectives and should be personalized based on the user. Figure 2, also developed based on the literature review, shows that about sixty-five percent of the work using a humanoid robot within the healthcare domain has been directed towards managing autism.

Application of humanoid robots in the field of education has also been promising. Studies have shown the positive influence of robots in education; moreover, humanoid robots have acted as a tour guide within the college campus [56]. Figure 3, developed based on the literature review findings, shows the most analyzed aspects of humanoids in education. From the literature review, it was found that application of a humanoid robot-enhanced the interaction of participants within a classroom setting, however [34], [36], [38], [39], no significant increase in learning rate were observed. The study conducted by Barakova in 2015 focused on analyzing the effect of using collocated robot and simulation on a student's enthusiasm. During the experiment with both robot and simulator, performance such a dropout rate, class attendance, task completion rate, creative thinking, and social impact were measured. All the measures were higher when working with a humanoid robot than that of a simulator; moreover, according to the questionnaire developed by Marina and Freidin in 2014 in their study, teacher, and students both preferred usage of a humanoid robot in the classroom setting. The study measured anxiety, attitude, adaptability, trust, and other, using a questionnaire [35], [38]. The study also observed that factors such as social presence and social influence are not relevant to determine a humanoid robot's acceptance [35].

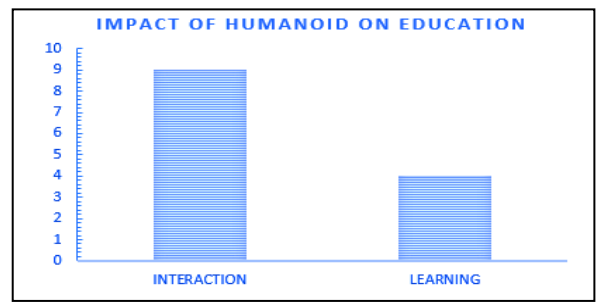

**Figure 3:** Humanoid's Impact on Education.

Moreover, students enjoyed and preferred the presence of humanoid robot in class [52][53]. Few studies have shown significant impact on the learning ability of the student using humanoid robots, but the presence of humanoid robot ensures the higher class attendance, promotes creative thinking [36] and increases subject curiosity [31]. Table 3 below shows the user's requirement from a typical humanoid robot specifically in the domain of private healthcare.

| Tasks | Percentage (Users asked for these features) |
| --- | --- |
| **Measuring BP, Body Temperature, Pulse, and Heartbeat irregularities. Sending the report to the doctor.** | 30.43% |
| **Connecting with a family doctor remotely** | 30.43% |
| **Medication reminder** | 30.07% |
| **Providing remote video connection for the doctors** | 28.97% |
| **Detecting health trouble, people lying (heart-attack) and call emergency help** | 27.87% |
| **Communicating with patient-doctor from time to time to ensure everything is ok** | 27.13% |
| **Ensure medications are taken at the proper dose** | 27.13% |
| **Reminding doctors to take care of the patient** | 26.03% |
| **Helping in speech therapy** | 26.03% |
| **Improving cognitive disability of the patient** | 25.67% |
| **Assisting in occupational therapy** | 25.30% |
| **Helping in baby care management** | 24.57% |
| **Helping in wound management and tube feeding** | 24.20% |
| **Monitoring injuries** | 23.83% |
| **Entertaining the patient** | 23.47% |
| **Assisting in mental therapy** | 22.73% |
| **Assisting in social skill and autism therapy** | 22.00% |

**Table 3:** Tasks for Home Healthcare Humanoid Robots.

## 3. Conclusion

The users have appreciated the role of humanoid robots in the field of healthcare and education. Contrastingly, people's attitude towards social or assistive robots varies significantly. Children and elderly users prefer robots and have less resistance towards the application of humanoid robots than that of middle-aged users. Trust and acceptance of humanoid robots were affected by its appearance, gaze, and functionality. According to a survey by Alaiad, people felt that using humanoid might be a threat to their privacy [54]. Humanoid robots were preferred more than general assistive robots [10], even people gave more attention to humanoid robots which are user-friendly [10]. Adult female users' trust decreased when the robots constantly gazed at them [18]; however, users with lower confidence had more trust towards the robot [18]. Humanoid robots were trusted with their functional knowledge such as weight, size, color, and other quantitative measures, but users did not trust on social and logical knowledge such good, bad, and other qualitative measures given by the robot [11]. People also believe that humanoid robots take undesirable actions intentionally and are more prone to make an error [20]. Unlike adult users, children were not concerned about the robot's utility. They enjoyed the company of humanoid robot [43] and treated them as a friend [42]. Children were willing and able to interact with the humanoid robot easily [55]. Old males had high concern about a robot's functionality [5] more than adolescent or females. Users want the humanoid robots to measure blood pressure, body temperature, connecting with a doctor remotely, reminding tasks, entertain, helping in the baby care management [54], lifting heavy, detect fall, control home appliance, housekeeping, making a phone call [45] and many other things.

The application of humanoid robots is crucial in the field of healthcare, education and as a social robot. In all these domains, most of the research is focused on the effect of using the humanoid robot, user acceptance and trust on the robot. It can be observed from the existing literature that the opinion of the people towards the use of humanoid robot varies from individual to individual. A user of different age, gender, and health condition perceive the usage and importance of humanoid robots differently.  The appearance of the robot was found to be a crucial factor affecting user acceptance and trust; moreover, robot functionality and gaze also changed user preference towards humanoid robot application.

Although many studies have been conducted to analyze the usage and effectiveness of humanoid robots, we still need more research in this field. As future research, the influence of incorporating a humanoid robot into the Project Leonard [56] as an assistant to care managers to enhance disease and care management effect must be addressed. Moreover, application of machine learning algorithms has been successfully able to assist doctors in diagnosing diseases [57]–[61], a humanoid robot programmed to do the same might have a more significant positive impact on a patient's health.

**Key Points**

- The Social connectivity of individual influences their perspective towards the use of a humanoid robot. Lonely individuals, especially autistic children and older adults prefer an assistive humanoid robot.
- The appearance of the robot influences user acceptance and trust. Users irrespective of their age, gender, social and health status preferred robots with human-like appearance.
- Application of humanoid robot to treat autism among children has been useful.
- Trust on a humanoid robot depends on the criticality of the situation a user finds herself or himself. Trust factor increases with a decrease in the confidence of a user in any quantitative decision-making situation.
- Humanoid robot in the education domain has been effective in increasing student participation.

> **Citation:** Avishek Choudhury, Huiyang Li, Christopher M Greene, Sunanda Perumalla. Humanoid Robot-Application and Influence. Archives of Clinical and Biomedical Research 2 (2018): 183-187.